\documentclass{article}
\usepackage{spconf,amsmath,graphicx}

\usepackage{multirow}
\usepackage{graphicx}
\usepackage{amsmath}
\usepackage{pifont}
\usepackage{blindtext}
\usepackage{hyperref}
\hypersetup{
    colorlinks=true,
    linkcolor=blue,
    filecolor=magenta,      
    urlcolor=cyan,
    pdftitle={Overleaf Example},
    pdfpagemode=FullScreen,
    }  
\usepackage{subcaption}
\usepackage[table]{xcolor}
\usepackage{booktabs}
\definecolor{lightgreen}{RGB}{200, 255, 200}
\definecolor{mediumgreen}{RGB}{100, 200, 100}
\definecolor{darkgreen}{RGB}{50, 150, 50}

\usepackage{tablefootnote}
\usepackage{gensymb}

\usepackage{enumitem}
\setlist{nosep, leftmargin=14pt}

\usepackage{mwe} 

\title{Non-Invasive 3D Wound Measurement with RGB-D Imaging}
\name{
\begin{tabular}{c}
Lena Harkämper$^{a,c}$, Leo Lebrat$^{b, c}$, David Ahmedt-Aristizabal$^{b, c}$, Olivier Salvado$^{b}$, \\
\textit{Mattias Heinrich$^{a}$,  Rodrigo Santa Cruz$^{b}$}
\end{tabular}
}

\address{$^{a}$ Institute of Medical Informatics, University of Lübeck, Lübeck, Germany \\
    $^{b}$ Queensland University of Technology, Brisbane, Australia \\
    $^{c}$ Imaging and Computer Vision Group, CSIRO Data61, Australia \\
\tt\normalsize lena.harkaemper@student.uni-luebeck.de,  r2.santacruz@qut.edu.au\\
}
\begin{document}
%
\maketitle
\begin{abstract}
Chronic wound monitoring and management require accurate and efficient wound measurement methods. 
This paper presents a fast, non-invasive 3D wound measurement algorithm based on RGB-D imaging. The method combines RGB-D odometry with B-spline surface reconstruction to generate detailed 3D wound meshes, enabling automatic computation of clinically relevant wound measurements such as perimeter, surface area, and dimensions.
We evaluated our system on realistic silicone wound phantoms and measured sub-millimetre 3D reconstruction accuracy compared with high-resolution ground-truth scans. The extracted measurements demonstrated low variability across repeated captures and strong agreement with manual assessments. The proposed pipeline also outperformed a state-of-the-art object-centric RGB-D reconstruction method while maintaining runtimes suitable for real-time clinical deployment. 
Our approach offers a promising tool for automated wound assessment in both clinical and remote healthcare settings.
\end{abstract}
\section{Introduction}
\label{sec:intro}

\begin{figure*}[htb]
  \centering
  \begin{minipage}[b]{0.48\linewidth}
    \centering
    \includegraphics[width=4.7cm]{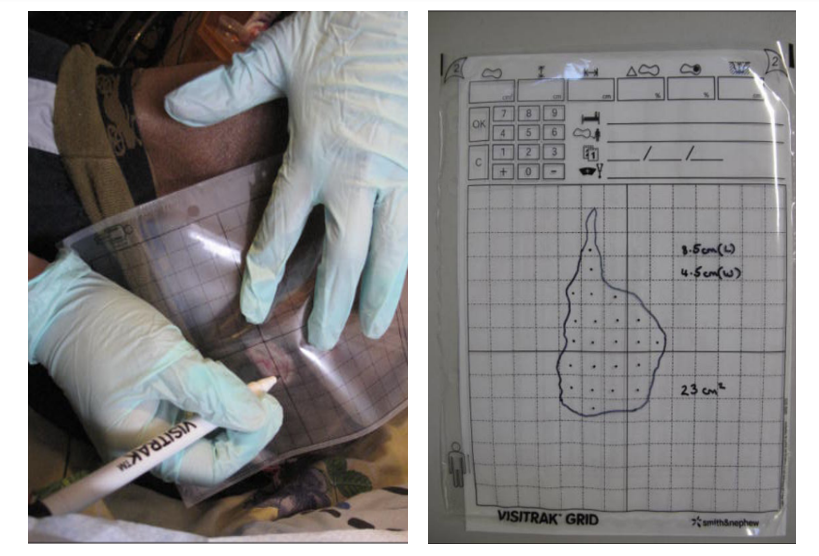}
    \subcaption{Manual wound tracing and surface area measurement~\cite{Elizabeth2015}.}
    \label{subfig:Manual_measurements}
  \end{minipage}
    ~
  \begin{minipage}[b]{0.48\linewidth}
    \centering
    \includegraphics[width=6.2cm]{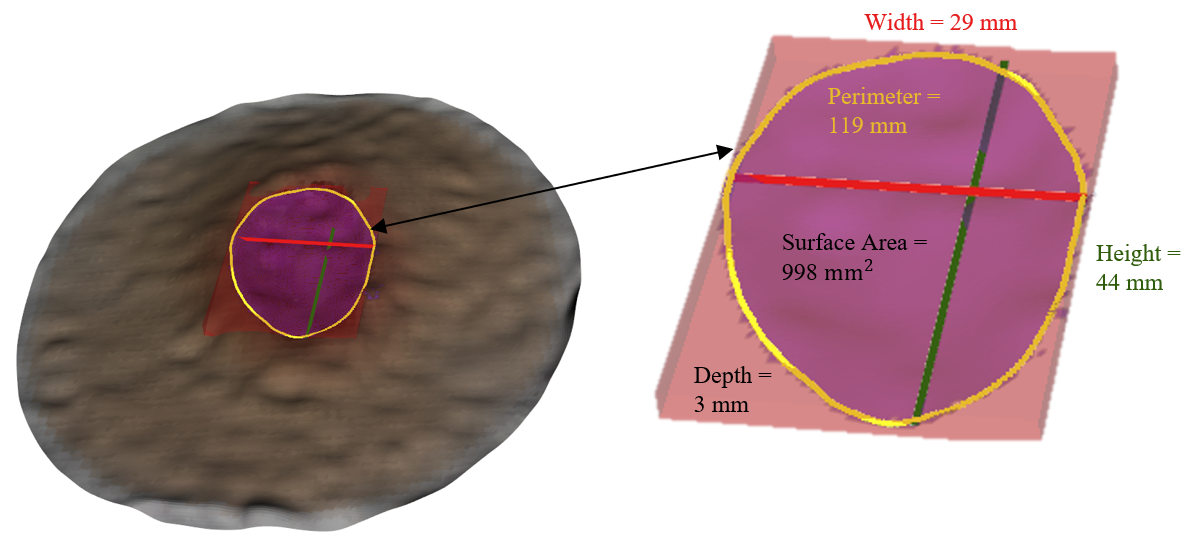}
    \subcaption{Segmented mesh (left) and automatic 3D wound bed measurements (right).}
    \label{subfig:Automatic_measurements}
  \end{minipage}
  \caption{Comparison of wound measurement methods: (a) 
  Manual tracing requires potentially harmful or invasive contact and is prone to clinician variability;
  (b) Our 3D approach enables non-invasive, accurate, reproducible wound assessment.}
  \label{fig:Motivational_figure}
\end{figure*}

Chronic wounds represent a significant health and economic concern worldwide~\cite{Sharma2024}. Adequate wound care requires accurate monitoring of wound measurements. In clinical practice, these measurements are conducted manually by healthcare professionals, by delineating wound outlines on transparent grids~\cite{Reifs2023} (Fig.~\ref{subfig:Manual_measurements}). 
This approach is invasive, time-consuming, and prone to inconsistencies due to practitioner experience, task ambiguity, and subjective interpretation.

Automated non-invasive wound measurement using cameras offers a cost-effective alternative, particularly in rural areas with limited access to specialists~\cite{Anisuzzaman2022}.
While commercial solutions~\cite{smartheal} rely on image-based measurements, 3D reconstruction provides angle-independent assessment and provides additional metrics (e.g., wound depth). Two main strategies exist for 3D wound measurement: multiview RGB reconstruction methods~\cite{Chierchia2024,chierchia2025wound3dassist} (such as photogrammetry and neural rendering) and RGB-D reconstruction using depth sensors. 
The former can achieve high accuracy but is sensitive to reflections and motion artifacts common in wound imaging, requires the placement of markers for scale recovery, and typically involves long processing times, limiting its clinical applicability.
In contrast, RGB-D methods offer faster results; however, existing approaches depend on GPU hardware~\cite{Filko2016, Filko2018}, have limited capture setups, and are frequently evaluated on wounds exhibiting simple geometries~\cite{Zhang2023}.

In this work, we propose a fast and accurate 3D wound measurement system using a commercial RGB-D camera and combining RGB-D odometry with B-spline surface reconstruction. Specifically, our method uses the Intel® RealSense™ D435\footnote{https://www.intelrealsense.com/depth-camera-d435/} to capture color point clouds from multiple viewpoints. 
These point clouds are aligned via relative camera motion estimation and fused into a smooth, consistent 3D wound model using a robust surface meshing algorithm. Clinically relevant metrics of the wound bed (i.e., the exposed viable tissue, excluding necrotic tissue and healthy surrounding skin) are automatically extracted, including perimeter, surface area and dimensions (Fig.~\ref{subfig:Automatic_measurements}).

We evaluate our approach in terms of both 3D reconstruction accuracy and measurement repeatability. Accuracy is assessed against high-resolution ground-truth scans of realistic silicone wound phantoms, which provide a stable, realistic, and non-disruptive environment for testing and comparing the proposed methods. Repeatability is evaluated by comparing automated results with manual measurements on the same phantoms. Our system achieves sub-millimeter reconstruction accuracy and yields wound measurements with $2.24\times$ higher consistency than manual assessments, representing an important first step toward implementing a portable, long-term chronic wound monitoring solution for telehealth and community nursing.

\section{Methods and materials}

\vspace{-6pt}
\subsection{Point cloud generation} \label{subsection:point_cloud_generation}
Our input consists of synchronized color and depth video streams, which are aligned and resampled to frames of $640 \times 480$ pixels using the RealSense SDK 2.0\footnote{https://github.com/IntelRealSense/librealsense}. 
For each frame, a point cloud is generated by back-projecting pixel coordinates into 3D space using the calibrated camera intrinsic matrix 
$\mathbf{K}$ and the corresponding depth map~\cite{Hartley2003}. Specifically, given a 2D pixel coordinate $(x, y)$ and its corresponding depth value $Z$, the 3D point $(X, Y, Z)$ in the camera coordinate system is computed as:


\begin{equation}
\begin{bmatrix} X \; Y \; Z \end{bmatrix}^\mathsf{T}
= Z\,\mathbf{K}^{-1}
\begin{bmatrix} x \; y \; 1 \end{bmatrix}^\mathsf{T}.
\label{eq:2D_to_3D_K}
\end{equation}

\vspace{-6pt}
\subsection{Point cloud registration}
\label{subsection:point_cloud_registration}
To create a unified 3D reconstruction, we merge the point clouds of multiple RGB-D frames. We select four evenly spaced frames and perform pairwise alignment to the first frame, for which we explore two methods:


\noindent\textbf{RGB-D Odometry-Based Alignment:} 
Given a source frame $(I_s, D_s)$ and a target frame $(I_t, D_t)$, RGB-D odometry aims to estimate a rigid transformation $\mathbf{T} \in \text{SE}(3)$ that best aligns with the target. This is achieved by minimizing a weighted combination of photometric and geometric errors:
\begin{align}
E(\mathbf{T}) =\ & (1 - \lambda) \sum_{\mathbf{p}} \left(I_t(\mathbf{p}') - I_s(\mathbf{p})\right)^2\nonumber \\ 
& + \lambda \sum_{\mathbf{p}} \left(D_t(\mathbf{p}') - \hat{D}_t(\mathbf{p}, D_s(\mathbf{p}), \mathbf{T})\right)^2.    
\end{align}

Here, $\mathbf{p}$ is a source pixel and $\mathbf{p}'$ its projection in the target image via $\mathbf{T}$. The first term measures the intensity difference between corresponding pixels, while the second compares the depth map value at the projected location in the target frame, $D_t(\mathbf{p}')$, with the expected depth $\hat{D}_t(\mathbf{p}, D_s(\mathbf{p}), \mathbf{T})$, computed as the Z-component of the transformed 3D point lifted from the source frame using Equation~\ref{eq:2D_to_3D_K}. 
The parameter $\lambda$ balances the photometric and geometric contributions. We use Open3D's implementation of the RGB-D odometry algorithm by Park et al.~\cite{Park2017}. 


\noindent\textbf{ArUco Marker-Based Frame Alignment: }
An alternative is to estimate the transformation between two RGB-D frames using fiducial markers. We placed two ArUco markers  (side length 1.3 cm) on opposite sides of the wound as consistent landmarks between frames.
For each frame, we detect the 2D coordinates of the marker corners using OpenCV’s ArUco module~\cite{Ramirez2018}, then lift them to 3D using Equation~\ref{eq:2D_to_3D_K}. This yields a correspondence set $\mathcal{K} = { (\mathbf{p}_i, \mathbf{q}_i) }_{i=1}^{8}$, where $\mathbf{p}_i$ and $\mathbf{q}_i$ are the 3D positions of the $i$-th marker corner in the source and target frames, respectively. A rigid transformation $\mathbf{T}$ is estimated by minimizing the point-to-point registration objective:

\begin{equation}
E(\mathbf{T}) = \sum_{(\mathbf{p}_i, \mathbf{q}_i) \in \mathcal{K}} \left\lVert \mathbf{T} \mathbf{p}_i -\mathbf{q}_i \right\rVert^2_2.
\end{equation}

\begin{figure*}[t!]
  \centering
  \begin{minipage}[b]{0.3\linewidth}
    \centering
    \includegraphics[height=2.7cm]{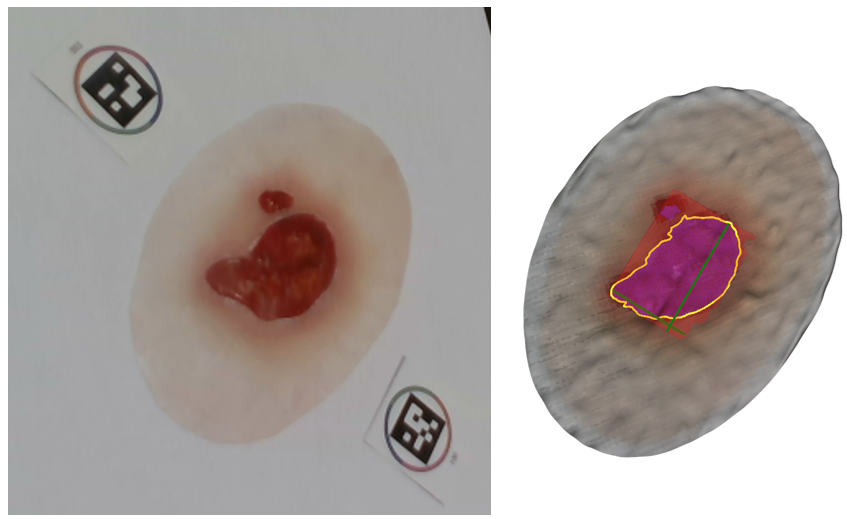}    
    \label{subfig:example_reconstructions_PIS3}
  \end{minipage}
    ~
  \begin{minipage}[b]{0.3\linewidth}
    \centering
    \includegraphics[height=2.7cm]{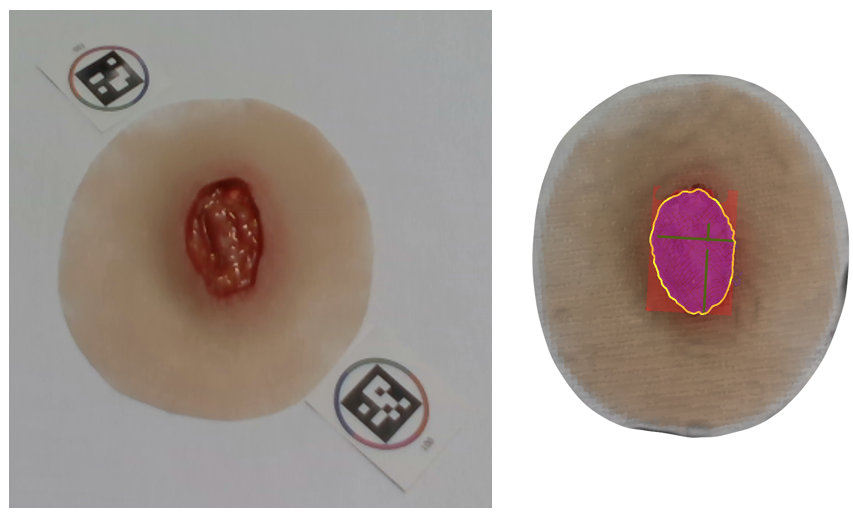}    
    \label{subfig:example_reconstructions_PIS4}
  \end{minipage}
    ~
  \begin{minipage}[b]{0.3\linewidth}
    \centering
    \includegraphics[height=2.7cm]{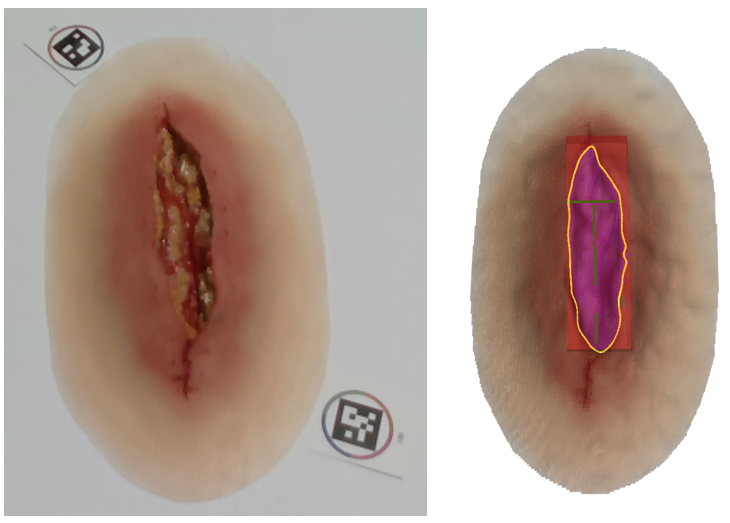}    
    \label{subfig:example_reconstructions_SD}
  \end{minipage}
  \caption{Visual comparison of silicone wound phantoms and their 3D reconstructions for three wound types (PIS3, PIS4, SD).}
  \label{fig:Reconstruction_examples}
\end{figure*}

\vspace{-6pt}
\subsection{Meshing algorithms}
\label{subsection:meshing_algorithms}
We evaluate several meshing algorithms to convert the merged point cloud into a triangle mesh. These include triangulation-based methods (Alpha Shapes~\cite{Edelsbrunner1983}, Ball Pivoting~\cite{Bernardini1999}), implicit surface reconstruction (Poisson~\cite{Kazhdan2006}), parametric surface approximation using B-spline surface fitting as proposed by Morwald et al.~\cite{Morwald2016} and implemented in the Point Cloud Library (PCL), and a neural implicit method (DiGS~\cite{Ben2022}). 

\subsection{3D Wound Measurements}
\label{subsection:wound_measurements}
\vspace{-6pt}
To compute clinically relevant wound bed measurements, we first segment the wound bed in the RGB images and map the 2D masks to the reconstructed wound mesh, from which the measurements are derived. Specifically:\\
\noindent\textbf{2D Wound Bed Segmentation:}
We segment the wound bed in the RGB frames using a deep neural network based on the SegFormer-B5 architecture (MiT-B5). The model was trained on a proprietary dataset with expert-annotated labels to produce per-pixel predictions, classifying each pixel as wound bed or background~\cite{chierchia2025wound3dassist}. \\
\noindent\textbf{Mesh Segmentation via 2D-to-3D Label Transfer:} 
To segment the mesh, we transfer labels from segmented RGB-D frames to the mesh via KNN label assignment. Each RGB-D frame is lifted to a point cloud, and each mesh vertex receives the majority label of its K nearest neighbors (Euclidean distance) in these point clouds. Aggregating multiple frames in 3D reduces 2D misclassifications, yielding a more robust segmentation and reliable measurements. \\
\noindent\textbf{3D Wound Bed Measurements:} 
From the segmented mesh, we extract the largest connected region of faces labeled as wound bed. Its open edges define the wound boundary, which is merged into a single closed loop and smoothed with a Savitzky–Golay filter \cite{Savitzky1964}. From this boundary, we compute:
\textit{1) Perimeter:} by fitting a 3D B-spline curve through the boundary vertices and computing its arc length; 
\textit{2) Surface Area:} by adding the areas of the triangular faces within the wound bed boundary;
\textit{3) Dimensions:} height, width, and depth as the three principal extents—ordered from largest to smallest—of the minimal volume bounding box enclosing the wound bed.



\section{Experiments}

\vspace{-6pt}
\subsection{Dataset}
We evaluated our method using three distinct types of realistic silicon wound phantoms provided by 
TraumaSIM\footnote{\url{https://traumasim.com.au/} (a company specializing in realistic simulations for healthcare training)}(Fig.~\ref{fig:Reconstruction_examples}): Pressure Injury Stage 3 (PIS3), Pressure Injury Stage 4 (PIS4), and Surgical Wound Dehiscence (SD). These phantoms offer a stable and repeatable testing environment while closely resembling real wounds.
High-resolution 3D ground-truth point clouds were acquired using a Zivid 2 M70 structured-light scanner. RGB-D videos were captured using an Intel® RealSense™ D435 camera\footnote{Initial calibration was performed once using the Intel® RealSense™ Dynamic Calibration API \url{https://dev.intelrealsense.com/docs/calibration}} under standard indoor ambient lighting, and RealSense™ spatial, temporal, and hole filling filters were applied.
To capture the complete wound geometry, the camera was moved in arcs at a 40–50 cm  distance, covering ±30\degree vertically and horizontally. Four evenly spaced frames were selected from each sequence for a global view.



\begin{table}[!t]
\centering
\footnotesize
\caption{
Point cloud 3D reconstruction metrics. 
}
\label{tab:Accuracy_Evaluation_Registrations}
\resizebox{\columnwidth}{!}{
\begin{tabular}{c r c c c c }
\toprule
\textbf{Wound} & \textbf{Method} & \textbf{AD} $\downarrow$ & \textbf{HD} $\downarrow$ & $\textbf{HD}_{90}$ $\downarrow$ & \textbf{t (s)} \\
\hline
\multirow{3}{*}{PIS3}
    & Single frame & 0.541 & \cellcolor{mediumgreen} 2.883 & \cellcolor{mediumgreen} 1.079 & - \\
    & Marker tracking & \cellcolor{lightgreen} 0.457 & 3.676 & \cellcolor{lightgreen} 1.324 & \cellcolor{mediumgreen} 5 \\
    & RGB-D odometry & \cellcolor{mediumgreen} 0.409 & \cellcolor{lightgreen} 3.564 & 1.375 & \cellcolor{lightgreen} 11 \\
\hline
\multirow{3}{*}{PIS4}
    & Single frame & 0.474 & 3.863 & \cellcolor{mediumgreen} 0.959 & -  \\
    & Marker tracking & \cellcolor{lightgreen} 0.460 & \cellcolor{lightgreen} 3.513 & 1.262 & \cellcolor{mediumgreen} 4  \\
    & RGB-D odometry & \cellcolor{mediumgreen} 0.436 & \cellcolor{mediumgreen} 3.751 & \cellcolor{lightgreen} 1.221 & \cellcolor{lightgreen} 11 \\
\hline
\multirow{3}{*}{SD}
    & Single frame & \cellcolor{lightgreen} 0.605 & 9.207 & \cellcolor{lightgreen} 1.680 & - \\
    & Marker tracking & 0.752 & \cellcolor{lightgreen} 8.404 & 1.734 & \cellcolor{mediumgreen} 5 \\
    & RGB-D odometry & \cellcolor{mediumgreen} 0.502 & \cellcolor{mediumgreen} 6.392 & \cellcolor{mediumgreen} 1.434 & \cellcolor{lightgreen} 12 \\ 
\bottomrule
\multicolumn{6}{p{150pt}}
{
$t$: registration time.
}
\end{tabular}
}
\end{table}

\begin{table*}[!th]
\centering
\footnotesize
\caption{Average clinical measurements and (max/mean) pairwise differences, in millimeters, of repeated manual and 3D measurements from B-spline odometry-based reconstructions.
}
\label{tab:Clinical_Measurements}
\begin{tabular}{c r c c c c c}
\toprule
\textbf{Wound} & \textbf{Method} & \textbf{Perimeter [mm]} & \textbf{Surface Area [mm$^2$]} & \textbf{Height [mm]} & \textbf{Width [mm]} & \textbf{Depth [mm]} \\
\hline
\multirow{2}{*}{PIS3} 
    & Manual        & 159.62 ($\textbf{6.70 / 3.22}$) &  1291.00 ($116.0 / 64.00$)  &  53.80 ($\textbf{4.00 / 2.20}$) & 31.60 ($\textbf{3.00 / 1.40}$) & 3.20 ($3.00 / 1.60$) \\
    & B-spline       & 140.62 ($11.46 / 5.80$)      &  1161.13 ($\textbf{60.58 / 29.20}$) &  47.02 ($5.72 / 2.51$)  & 34.69 ($5.16 / 2.24$) & 3.67 ($\textbf{2.31 / 1.14}$) \\
\hline
\multirow{2}{*}{PIS4}
    & Manual        & 129.98 ($11.20 / 4.72$)         & 980.60 ($109.00 / 46.20$)    &  41.00 ( $11.00 / 6.40$ ) & 27.20 ($4.00 / 1.80$) & 6.00 ($3.00 / 1.40$) \\
    & B-spline     &  118.75 ($\textbf{0.72 / 0.37}$) &  982.75 ($\textbf{24.70 / 12.15}$)  &  42.69 ($\textbf{3.66 / 2.07}$) & 30.95 ($\textbf{3.15 / 1.56}$) & 2.94 ($\textbf{0.33 / 0.15}$)  \\
\hline
\multirow{2}{*}{SD} 
    & Manual        & 326.92 ($65.30 / 28.54$)        & 2617.20 ($466.00 / 222.00$)   &  133.20 ($16.00 / 6.80$) & 26.90 ($2.50 / 1.40$)  & 20.80 ($8.00 / 4.00$) \\
    & B-spline       &  247.43 ($\textbf{5.49 / 2.64}$)     &  3016.59 ($\textbf{92.24 / 46.19}$)  &  112.99 ($\textbf{4.94 / 2.56}$) & 32.48 ($\textbf{0.29 / 0.13}$) & 12.54 ($\textbf{0.86 / 0.46}$) \\
\bottomrule
\end{tabular}
\end{table*}

\vspace{-6pt}
\subsection{3D reconstruction accuracy}
To assess reconstruction accuracy, we compare point clouds and meshes against ground-truth scans using standard 3D reconstruction metrics: Average Distance (AD), Hausdorff Distance at the 100th (HD) and 90th (HD$_{90}$) percentiles (in mm), and Normal Consistency (NC), ranging from 0 (orthogonal normals) to 1 (perfectly aligned normals)~\cite{Chierchia2024}. All reconstructions are first aligned to the ground truth via manual initialization, followed by point-to-point ICP refinement 
using point cloud processing software. 
After alignment, the data is tightly cropped around the wound region to ensure that evaluation focuses solely on the area of interest. For meshes specifically, the metrics are computed on a uniformly sampled point cloud.

In Table~\ref{tab:Accuracy_Evaluation_Registrations}, we compare the accuracy of single-frame point clouds to dense reconstructions obtained by merging frame-wise point clouds using either ArUco marker tracking or RGB-D odometry. While single-frame reconstructions already achieve sub-millimeter accuracy, they often miss fine details in regions with limited viewpoint coverage. Merging multiple frames enhances reconstruction quality by integrating complementary geometry from different perspectives. 
RGB-D odometry consistently delivers the best or near-best performance across most metrics, particularly for more complex geometries like SD. We hypothesize that this improved performance is due to the additional photometric term, which helps handle noise in depth maps. In contrast, marker tracking relies solely on depth readings at the marker corners, which can be inaccurate. All methods run within seconds, supporting their feasibility for clinical deployment. 

\begin{table}[!t]
\centering\
\footnotesize
\caption{
Mesh 3D reconstruction metrics.
}
\label{tab:Accuracy_Evaluation_Meshing}
\resizebox{\columnwidth}{!}{
\begin{tabular}{c r c c c c c }
\toprule
\textbf{Wound} & \textbf{Meshing method}& \textbf{AD} $\downarrow$ & \textbf{HD} $\downarrow$ & $\textbf{HD}_{90}$ $\downarrow$ & \textbf{NC} $\uparrow$ & \textbf{t (s)} \\
\hline
\multirow{7}{*}{PIS3}
    & B-spline & \cellcolor{mediumgreen} 0.343 & \cellcolor{mediumgreen} 2.367 & \cellcolor{mediumgreen} 0.796 & \cellcolor{mediumgreen} 0.952 & 11+16 \\
    & DiGS& 0.418 & 3.388 & 1.274 & 0.942 & 11+801 \\
    & Alpha shapes & 1.012 & 3.724 & 1.811 & 0.918 &  11+3  \\
    & BPA & 0.584 & 3.550 & 1.518 & 0.778 & \cellcolor{lightgreen} 11+2  \\
    & Poisson & 0.424 & \cellcolor{lightgreen} 2.981 & \cellcolor{lightgreen} 1.040 & 0.822 & \cellcolor{mediumgreen} 11+1 \\
    & BundleSDF & \cellcolor{lightgreen} 0.367  & 19.561 & 0.561 & \cellcolor{lightgreen} 0.945 & 346 \\
\hline
\multirow{7}{*}{PIS4}
    & B-spline & \cellcolor{lightgreen} 0.426 & 3.808 & \cellcolor{lightgreen} 0.997 & \cellcolor{mediumgreen} 0.953 & 11+16 \\
    & DiGS& 0.452 & 4.594 & 1.344 & \cellcolor{lightgreen} 0.946 & 11+781 \\
    & Alpha shapes & 0.810 & 3.765 & 1.699 & 0.926 & 11+6 \\
    & BPA & 0.494 & \cellcolor{lightgreen} 3.715 & 1.424 & 0.790 & \cellcolor{lightgreen} 11+2 \\
    & Poisson & \cellcolor{mediumgreen} 0.377 & \cellcolor{mediumgreen} 3.599 & \cellcolor{mediumgreen} 0.913 & 0.844 & \cellcolor{mediumgreen} 11+1 \\
    & BundleSDF & 0.639  & 15.312 & 1.463 & 0.932 & 333 \\
\hline
\multirow{7}{*}{SD}
    & B-spline & \cellcolor{lightgreen} 0.465 & 7.750 & 1.271 & 0.774 & 12+38 \\
    & DiGS & 0.972 & 11.359 & 4.525 & \cellcolor{lightgreen} 0.799 & 12+727 \\
    & Alpha shapes & 0.687 & \cellcolor{mediumgreen} 6.395 & 1.481 & \cellcolor{mediumgreen} 0.822& \cellcolor{lightgreen} 12+10   \\
    & BPA & 0.493 & \cellcolor{lightgreen} 6.595 & \cellcolor{lightgreen} 1.236 & 0.712 &  \cellcolor{mediumgreen} 12+3  \\
    & Poisson & \cellcolor{mediumgreen} 0.450 & 7.282 & \cellcolor{mediumgreen} 1.100 & 0.721 & 12+20 \\
    & BundleSDF & 0.891  & 7.574 & 2.040 & 0.818 & 451 \\
\bottomrule
\multicolumn{7}{p{200pt}}
{
$t$: total computation time (registration + meshing). 
}

\end{tabular}
}
\end{table}

Table~\ref{tab:Accuracy_Evaluation_Meshing} compares the mesh reconstruction accuracy of five meshing methods applied to point clouds obtained with RGB-D odometry-based multi-frame 3D reconstruction. Additionally, we include results obtained with BundleSDF~\cite{BundleSDF} as a SOTA reference for object-centric 3D reconstruction from RGB-D data. Across all wound types, the B-spline and Poisson methods generally achieve the best accuracy scores, with B-spline exhibiting the lowest AD and HD values for PIS3 and strong performance in PIS4 and SD as well. Poisson follows closely, excelling in PIS4, where it delivers the best performance across all geometric metrics. DiGS yields good NC values but suffers from long runtimes. 
BundleSDF provides good results but is surpassed by the proposed techniques in both accuracy and runtime, taking several minutes for some reconstructions. 
In summary, B-spline offers the best balance between accuracy and efficiency, which motivates its selection as the meshing method used in the following experiments.

\vspace{-6pt}
\subsection{Wound measurements reliability}
To assess the suitability of our approach for wound monitoring, we compare the consistency of wound measurements from our 3D reconstructions (Section~\ref{subsection:wound_measurements}) with a manual measurement method (Fig.~\ref{subfig:Manual_measurements}). 
Five 3D reconstructions per wound are generated using different sets of four RGB-D frames, RGB-D odometry, and B-spline meshing. For the manual method, five participants with similar backgrounds delineate the wound boundary on a transparent sheet placed over each silicone phantom. Width and height are measured with a ruler, and depth is estimated by using a standardized swab method. Perimeter and surface area are computed from images of the sheet using image processing techniques. 

Table~\ref{tab:Clinical_Measurements} reports the average clinical measurements for each wound type across the five reconstructions or five manual measurement attempts. Maximum and mean pairwise differences are shown in brackets to quantify variability.
The proposed approach shows lower variability than manual measurements for PIS4 and SD, particularly in perimeter and surface area, indicating greater repeatability and robustness. Manual assessments exhibit larger variation, likely due to their subjective nature, despite being performed in controlled conditions (no patient interaction or discomfort). For PIS3, the proposed method shows slightly higher variability due to the presence of two wound beds, which were inconsistently segmented across reconstructions. 
Wound healing guidelines specify that a decrease in wound surface area by 10–15\% per week is a meaningful indicator of therapeutic progress~\cite{Foltynski2021}. The proposed method exhibits mean surface area variability below 3\%, well below the clinically meaningful 10–15\% change, indicating adequate precision for detecting meaningful wound size changes over time.

\section{Conclusion}
This work presents a fast and accurate pipeline for non-invasive 3D wound measurement using an RGB-D camera and combining RGB-D odometry with B-spline surface reconstruction.
It achieves sub-millimeter accuracy, outperforms marker-based registration and SOTA method BundleSDF, and yields more consistent measurements than manual assessment with sufficient precision to detect clinically meaningful wound area changes. 
As future work, we plan to validate the technology with real patients in remote areas, which represents the next step toward deployment in telehealth settings. We will investigate the system’s robustness to common real-world challenges, such as patient motion artifacts (e.g., breathing).
Additionally, a sensitivity analysis will quantify the impact of errors in 2D segmentation masks on final 3D area measurements. Finally, protocol stress tests will evaluate performance under challenging acquisition conditions, such as varying working distances, angle coverage, lighting conditions, number of frames, and motion blur.

\section{Compliance with Ethical Standards}
\vspace{-5pt}
This study was performed in line with the principles of the Declaration of Helsinki. The experimental procedures involving human subjects described in this paper were approved by CSIRO Health and Medical Human Research Ethics Committee (CHMHREC).

{
    \bibliographystyle{IEEEbib}
    \bibliography{strings,refs}

@article{Sharma2024,
    author = {Aditya Sharma and others},
    title ={Burden of Chronic Nonhealing Wounds: An Overview of the Worldwide Humanistic and Economic Burden 
    to the Healthcare System},
    journal = {The International Journal of Lower Extremity Wounds},
    pages = {15347346241246339},
    year = {2024},
    doi = {10.1177/15347346241246339},
    note ={PMID: 38659348}}

@article{Reifs2023,
    title = {Clinical validation of computer vision and artificial intelligence algorithms for wound measurement and tissue classification in wound care},
    journal = {Informatics in Medicine Unlocked},
    volume = {37},
    pages = {101185},
    year = {2023},
    issn = {2352-9148},
    doi = {10.1016/j.imu.2023.101185},
    author = {David Reifs and others}
}

@article{Anisuzzaman2022,
   title={Image-Based Artificial Intelligence in Wound Assessment: A Systematic Review},
   volume={11},
   ISSN={2162-1934},
   DOI={10.1089/wound.2021.0091},
   number={12},
   journal={Advances in Wound Care},
   publisher={Mary Ann Liebert Inc},
   author={Anisuzzaman, D.M. and others},
   year={2022},
   pages={687–709} }

@inproceedings{Chierchia2024,
  title={{SALVE}: A 3D Reconstruction Benchmark of Wounds from Consumer-Grade Videos},
  author={Chierchia, Remi and others},
  booktitle={2025 IEEE/CVF Winter Conference on Applications of Computer Vision (WACV)},
  pages={4205--4214},
  year={2025},
  organization={IEEE}
}

@book{Hartley2003,
  title={Multiple View Geometry in Computer Vision},
  author={Hartley, Richard and Zisserman, Andrew},
  year={2003},
  publisher={Cambridge University Press},
  doi = {10.1017/CBO9780511811685}
}

@article{Savitzky1964,
    author = {Savitzky, Abraham and Golay, Marcel J. E.},
    title = {Smoothing and Differentiation of Data by Simplified Least Squares Procedures.},
    journal = {Analytical Chemistry},
    volume = {36},
    number = {8},
    pages = {1627-1639},
    year = {1964},
    doi = {10.1021/ac60214a047}
}

@INPROCEEDINGS{Park2017,
  author={Park, Jaesik and others},
  booktitle={2017 IEEE International Conference on Computer Vision (ICCV)}, 
  title={Colored Point Cloud Registration Revisited}, 
  year={2017},
  volume={},
  number={},
  pages={143-152},
  doi={10.1109/ICCV.2017.25}}

@article{Ramirez2018,
author = {Romero-Ramirez, Francisco and others},
year = {2018},
month = {06},
pages = {},
title = {Speeded Up Detection of Squared Fiducial Markers},
volume = {76},
journal = {Image and Vision Computing},
doi = {10.1016/j.imavis.2018.05.004}
}

@ARTICLE{Edelsbrunner1983,
  author={Herbert Edelsbrunner and others},
  journal={IEEE Transactions on Information Theory}, 
  title={On the shape of a set of points in the plane}, 
  year={1983},
  volume={29},
  number={4},
  pages={551-559},
  doi={10.1109/TIT.1983.1056714}}

@ARTICLE{Bernardini1999,
  author={Fausto Bernardini and others},
  journal={IEEE Transactions on Visualization and Computer Graphics}, 
  title={The ball-pivoting algorithm for surface reconstruction}, 
  year={1999},
  volume={5},
  number={4},
  pages={349-359},
  doi={10.1109/2945.817351}}

@inproceedings{Kazhdan2006,
author = {Kazhdan, Michael and others},
title = {Poisson surface reconstruction},
year = {2006},
isbn = {3905673363},
publisher = {Eurographics Association},
address = {Goslar, DEU},
booktitle = {Proceedings of the Fourth Eurographics Symposium on Geometry Processing},
pages = {61–70},
numpages = {10},
location = {Cagliari, Sardinia, Italy},
series = {SGP '06}
}

@article{Morwald2016,
title = {Modeling connected regions in arbitrary planar point clouds by robust B-spline approximation},
journal = {Robotics and Autonomous Systems},
volume = {76},
pages = {141-151},
year = {2016},
issn = {0921-8890},
doi = {10.1016/j.robot.2015.11.006},
author = {Thomas Mörwald and others},
}

@inproceedings{Ben2022,
  title={Digs: Divergence guided shape implicit neural representation for unoriented point clouds},
  author={Ben-Shabat, Yizhak and others},
  booktitle={Proceedings of the IEEE/CVF Conference on Computer Vision and Pattern Recognition},
  pages={19323--19332},
  year={2022}
}

@article{Elizabeth2015,
    author = {Nichols, Elizabeth},
    title = {Wound assessment part 1: how to measure a wound},
    journal = {Wound Essentials},
    year = {2015},
    volume={10},
    number={2},
    pages={51-54}
}

@ARTICLE{Zhang2023,
  author={Zhang, Peng and others},
  journal={IEEE Transactions on Instrumentation and Measurement}, 
  title={RGB-D Camera-Based Automatic Wound-Measurement System}, 
  year={2023},
  volume={72},
  number={},
  pages={1-11},
  doi={10.1109/TIM.2023.3265758}}

@article{Filko2018,
  title={Wound measurement by RGB-D camera},
  author={Filko, Damir and others},
  journal={Machine Vision and Applications},
  year={2018},
  volume={29},
  pages={633 - 654},
  url={https://api.semanticscholar.org/CorpusID:4473153}
}

@article{Filko2016,
title = {Detection, Reconstruction and Segmentation of Chronic Wounds Using Kinect v2 Sensor},
journal = {Procedia Computer Science},
volume = {90},
pages = {151-156},
year = {2016},
note = {20th Conference on Medical Image Understanding and Analysis (MIUA 2016)},
issn = {1877-0509},
doi = {https://doi.org/10.1016/j.procs.2016.07.022},
url = {https://www.sciencedirect.com/science/article/pii/S1877050916312005},
author = {Filko, Damir and others}}

@inproceedings{BundleSDF,
  title={Bundlesdf: Neural 6-dof tracking and 3d reconstruction of unknown objects},
  author={Wen, Bowen and others},
  booktitle={Proceedings of the IEEE/CVF Conference on Computer Vision and Pattern Recognition},
  pages={606--617},
  year={2023}
}

@misc{smartheal,
  title = {SmartHeal},
  howpublished = {\url{https://smartheal.org}},
  note = {Accessed: 2024-07-12}
}

@article{chierchia2025wound3dassist,
  title={Wound3DAssist: A Practical Framework for 3D Wound Assessment},
  author={Chierchia, Remi and others},
  journal={arXiv preprint arXiv:2508.17635},
  year={2025}
}

@article{Foltynski2021,
title = {Wound surface area measurement methods},
journal = {Biocybernetics and Biomedical Engineering},
volume = {41},
number = {4},
pages = {1454-1465},
year = {2021},
issn = {0208-5216},
doi = {https://doi.org/10.1016/j.bbe.2021.04.011},
author = {Piotr Foltynski and others}
}
}

\end{document}